\documentclass[11pt]{article}

\usepackage[final]{acl}
\usepackage{times}
\usepackage{latexsym}

\usepackage[T1]{fontenc}

\usepackage[utf8]{inputenc}

\usepackage{microtype}

\usepackage{inconsolata}

\usepackage{graphicx}
\usepackage{algorithm}
\usepackage{algorithmic}

\usepackage{newfloat}
\usepackage{listings}
\DeclareCaptionStyle{ruled}{labelfont=normalfont,labelsep=colon,strut=off} 
\floatstyle{ruled}
\newfloat{listing}{tb}{lst}{}
\floatname{listing}{Listing}
\usepackage{multirow}
\usepackage{booktabs}
\usepackage{graphicx}
\usepackage{subfig}
\usepackage{ragged2e}
\usepackage{array}
\usepackage[none]{hyphenat}
\usepackage{mathtools}
\usepackage{cleveref}

\newcolumntype{R}[1]{>{\RaggedLeft\arraybackslash}p{#1}}
\newcolumntype{L}[1]{>{\RaggedRight\arraybackslash}p{#1}}

\title{Surprisingly Fragile: Assessing and Addressing Prompt Instability in Multimodal Foundation Models}

\author{
Ian Stewart, Sameera Horawalavithana, Brendan Kennedy, Sai Munikoti, Karl Pazdernik \\
Pacific Northwest National Laboratory \\
Richland, WA 99354 \\
\texttt{\{ian.stewart,yasanka.horawalavithana,brendan.kennedy,sai.munikoti,karl.pazdernik\}@pnnl.gov}
}

\begin{document}

\maketitle
\begin{abstract}
Multimodal foundation models (MFMs) such as OFASys show the potential to unlock analysis of complex data such as images, videos, and audio data via text prompts alone.
However, their performance may suffer in the face of text input that differs even slightly from their training distribution, which is surprising considering the use of modality-specific data to ``ground'' the text input.
This study demonstrates that prompt instability is a major concern for MFMs, leading to a consistent drop in performance across all modalities, but that instability can be mitigated with additional training with augmented data.
We evaluate several methods for \emph{grounded prompt perturbation}, where we generate perturbations and filter based on similarity to text and/or modality data.
After re-training the models on the augmented data, we find improved accuracy and more stable performance on the perturbed test data regardless of perturbation condition, suggesting that the data augmentation strategy helps the models handle domain shifts more effectively.
In error analysis, we find consistent patterns of performance improvement across domains, suggesting that retraining on prompt perturbations tends to help general reasoning capabilities in MFMs.
\end{abstract}

\section{Introduction}

While text-only large language models (LLMs) are powerful in their own right, multimodal foundation models (MFMs) offer even more transformative capabilities such as image captioning and audio signal processing~\cite{li2024}.
Generalist models such as Unified-IO are trained to process a range of modalities including image, video, and audio signal, often using an LLM as a combined encoder for all input types after preprocessing ~\cite{lu2022}.
MFMs have the potential for major impact in downstream applications such as autonomous navigation~\cite{cui2024} and medical image analysis~\cite{moor2023}, provided that their output is consistent in the face of variable user input.


Despite their power, these models often show surprising fragility in the face of prompts that have slight differences from the expected syntax or semantics seen during training~\cite{newman2021}.
We show an example from a popular video QA dataset (iVQA) in Table~\ref{tab:example_prompt_perturb_response}, in which a person fixes a car engine and the model is asked to respond to the relevant prompt ``What mechanism is the person working on?''.
In response to prompt \emph{perturbations} (paraphrased prompts), a baseline model with no additional training responds with a variety of incorrect responses.
In contrast, a model that has been trained on prompt perturbations (``Joint similarity''; see Section~\ref{sec:data_sampling}) responds correctly regardless of phrasing.

\begin{table}[t!]
    \centering
    \scriptsize
    \begin{tabular}{p{3cm} L{1.5cm} L{2cm}}
    Data & Prompt                    
     & Answer                                                              \\ \toprule
     \includegraphics[width=3cm]{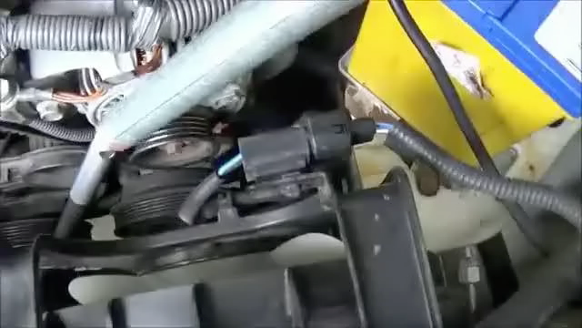} & What mechanism is the person working on? & The person in the video seems to be working on an engine mechanism. \\ \\ \toprule \toprule \\
     Perturbed prompt & Model response (no training) & Model response (trained on prompt perturbations) \\ \\ \midrule \\
     what project is the person working on? & home & The person in the video is working with a car engine. \\
    what technology is the person improving? & oil & The person in the video appears to be working on a car. \\
    what task is the person in charge of? & excavator & The person in charge of the vehicle seems to be working on the engine.
    \end{tabular}
    \caption{Unified-IO model responses to prompt perturbations on example data (iVQA video, first frame shown).}
    \label{tab:example_prompt_perturb_response}
\end{table}


This study addresses a central research question in model robustness: to what degree does re-training multimodal models on prompt perturbations, chosen by different selection strategies, improve robustness?
We investigate the relative stability of current state-of-the-art MFMs on several standard multimodal benchmark datasets and find the following:

\begin{itemize}
    \itemsep0em
    \item Modern MFMs show a severe drop in performance on the image domain, in response to perturbed prompts on standard QA datasets (\ref{sec:testing_perturbations}).
    \item Re-training MFMs on a wider variety of perturbed prompts, regardless of how the prompts are selected (\ref{sec:data_sampling}), improves performance across domains and reduces instability on other prompt perturbations, more than simply re-training on the original prompts (\ref{sec:training_perturbations}). 
    \item The perturbation-trained models show consistent improvement when the prompts target modality-specific information, e.g. prompts in the image modality that reference material properties (\ref{sec:error_analysis}).
\end{itemize}

Our work provides an automated method to help align MFMs to human expectations, such as consistency in response to input variation~\cite{zhao2024}.

\section{Related work}

\subsection{Multimodal foundation models}

Recent work has unified text-based foundation models with modality-specific processing to yield MFMs that can respond to input from a variety of modalities~\cite{li2024}.
The majority of such models, including Unified-IO and OFASys, first encode the text and multimodal data to a shared representation space, then decode the combined representation into the desired modality, often text (e.g. image QA requires encoding image + text data and decoding as a text response).
Some systems such as Next-GPT learn to perform any-to-any modality generation by training modality-specific decoders~\cite{wu2023}, which works reasonably well for more common modalities such as images.
For breadth, systems including ImageBind~\cite{girdhar2023} and OneLLM~\cite{han2023} use a universal encoder and learn from a wide variety of data including depth maps and point clouds, which may help the model's internal LLM achieve better cross-modality understanding.
To further improve performance, parameter-efficient methods such as prompt learning have helped to solve problems with missing data~\cite{lee2023} and cross-modality alignment~\cite{khattak2023}.
We identify a possible gap in the research space, i.e. model-agnostic methods to improve robustness, and test a variety of prompt perturbation methods to determine the relative impact on task performance.

\subsection{Model stability}

Many LLMs struggle with limited \emph{stability}: the response to a given input may vary widely depending on changes in input wording or semantics~\cite{newman2021}, even as subtle as adding extra punctuation~\cite{formento2023}.
This not only presents a negative user experience to end users, requiring them to iterate on the same prompt~\cite{kim2024}, but also opens vulnerability to attacks that can lead to model output of harmful information~\cite{schulhoff2023}.
One common solution to this problem is model regularization, i.e. forcing the model to learn general patterns in the data, and methods for regularization include enforcing sparsity in model architecture~\cite{li2023}, or more commonly augmenting training data with ``noised'' data points~\cite{chen2023}.
In the multimodal space, methods such as synonym replacement have shown to generate useful data points at inference time to reduce instability and improve overall performance~\cite{ge2023}.
We address the issue of stability through modality-grounded prompt perturbation and show that re-training on prompts, regardless of sampling strategy, can help protect models from performance drops on unseen prompts.

\section{Methods}

We first outline the methods used for sampling data-grounded prompts for further model training (\ref{sec:data_sampling}), identify the models trained using the augmented data (\ref{sec:models}), and explain the QA data used in the experiments (\ref{sec:data}).

\subsection{Prompt augmentation}
\label{sec:data_sampling}


Typical approaches to data augmentation include utilizing a variety of similar but distinct prompts to re-train the model, e.g. randomly changing the prompt instructions during training~\cite{lu2023}.
We test several simple methods for data augmentation to determine whether MFMs require modality-specific prompts or text-specific prompts to improve overall prompt stability.

For a given text prompt $t$, we generate a list of $N=10$ perturbations $\hat{T}$ with a general-purpose LLM,\footnote{Llama-2, accessed 1 Jan 2024: \url{https://huggingface.co/meta-llama/Llama-2-7b-chat-hf}.} using instructions to maximize diversity and semantic similarity to the original prompt.\footnote{Instruction: ``Write 10 new questions that are paraphrases of the first question. The new questions should be unique but should have the same meaning as the first question. First question:''}
We choose 10 as the number of perturbations to maximize prompt diversity while avoiding prompts that ``drift'' too far from the original prompt (e.g., generating 100 perturbations could yield highly divergent perturbations).
This full set of perturbations $\hat{T}$ is used in evaluation to determine post-training performance of the models on the perturbed prompt data.

We then compute the latent representation $X$ of text ($X_{t}$) and multimodal data ($X_{m}$) using the ImageBind model~\cite{girdhar2023}, an encoder trained to align latent representations of data from different modalities with contrastive learning, using text as the intermediate modality.
Finally, we select $k=3$ prompt perturbations from the set $\hat{T}$ with maximum cosine similarity between the prompt perturbation text and the data $X$.
In this step, we choose either $X_{t}$ for perturbations most similar to the text data (``text-similar sampling'') or $X_{m}$ for perturbations most similar to the modality data (``modality-similar sampling'').
As a control condition, we also sample randomly from the prompt perturbations (``random sampling'').
We choose 3 as the number of prompts per category to balance semantic diversity with a reasonable runtime for training and a need to avoid overfitting.

In addition to these simpler conditions for prompt perturbation, we test a method to balance similarity of the perturbed prompt to the original data and diversity among the prompts.
In the first step, we sample 1 prompt perturbation $\hat{t}$ from $\hat{T}$ based on Eq.~\ref{eqn:joint_sampling} similarity to original data (text and multimodal data).

\begin{equation}
\label{eqn:joint_sampling}
P(\hat{t}) \propto \text{JointSim}(\hat{t}) = sim(\hat{t}, X_{t}) + sim(\hat{t}, X_{m})
\end{equation}
For all subsequent prompt samples, in order to increase the diversity of the jointly sampled prompts, we normalize the similarity relative to the cumulative similarity of the original prompt to the set of already-sampled prompts ($\hat{T}'$) from prior iterations (Eq.~\ref{eqn:diversity_norm}). As each additional prompt perturbation $\hat{t}$ is sampled, it is added to $\hat{T}'$. 

\begin{equation}
\label{eqn:diversity_norm}
P(\hat{t}) \propto \frac{\text{JointSim}(\hat{t})}
{\frac{1}{|\hat{T}'|}\sum_{\hat{t}' \in \hat{T}'} sim(\hat{t}, \hat{t}')}
\end{equation}

This method (``Joint similarity'') yields prompt perturbations that are divergent from one another and also \emph{grounded} with respect to both text data $X_{t}$ and modality data $X_{m}$, thus decreasing the risk of re-training the model on augmented prompts that are irrelevant to the input context.

\begin{table}[t!]
\scriptsize
\centering
\begin{tabular}{L{2.3cm} L{2.3cm} L{2cm}}
\multicolumn{1}{L{2.3cm}}{Prompt} & \multicolumn{2}{l}{Data} \\ \toprule
\multicolumn{1}{L{2.3cm}}{What is the person holding?} & \multicolumn{2}{l}{\includegraphics[width=4cm]{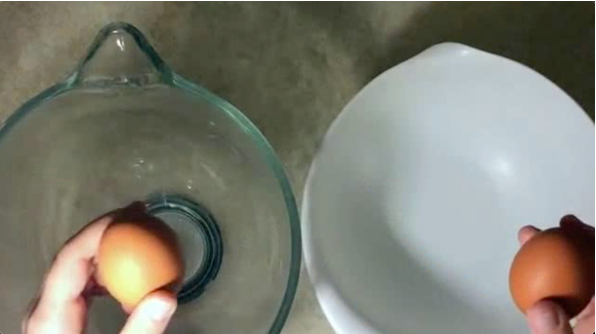}} \\ \midrule
\multicolumn{3}{l}{Prompt perturbations} \\ \midrule
\multicolumn{3}{l}{\begin{tabular}[c]{@{}l@{}}1. what is the person's grasping tool?\\ 2. what is the person's hold?\\ 3. what object is the person gripping?\\ 4. what is the person's holding device?\\ 5. what is the person's grasp?\\ 6. what item is the person clutching?\\ 7. what is the person's grip?\\ 8. what is the person's clutch?\\ 9. what object is the person grasping?	\\ 10. what is the person holding tightly?\end{tabular}} \\ \midrule
Text similarity & Modality similarity & Joint similarity \\ \midrule
\begin{tabular}[c]{@{}L{2.3cm}@{}}2. what is the person's hold?\\ 10. what is the person holding tightly?\\ 4. what is the person's holding device?\end{tabular} & \begin{tabular}[c]{@{}L{2.3cm}@{}}6. what item is the person clutching?\\ 7. what is the person's grip?\\ 5. what is the person's grasp?\end{tabular} & \begin{tabular}[c]{@{}L{2.3cm}@{}}8. what is the person's clutch?\\ 7. what is the person's grip?\\ 6. what item is the person clutching?\end{tabular}
\end{tabular}
\caption{Prompt perturbation sampling methods from~\ref{sec:data_sampling}, with example data and prompt (video frame from iVQA).}
\label{tab:prompt_sample_method}
\end{table}

We summarize the prompt sampling methods in Table~\ref{tab:prompt_sample_method}, where we show an example data point with the corresponding sampled prompt perturbations.


\subsection{Models}
\label{sec:models}

For our experiments, we tested image, video, and spoken audio data,\footnote{At time of writing, no open-source generalist models prove capable of using general-domain audio as input, only human speech audio.} which are three related but complementary data modalities that a MFM should reasonably be expected to handle.
Despite the proliferation of literature related to MFMs, relatively few models are available for open-source inference, and even fewer for fine-tuning on new datasets without extensive preprocessing.
We investigate the following models which have released open-source implementations including training code:

\begin{itemize}
    \item OFASys\footnote{Accessed 1 Jan 2024: \url{https://github.com/OFA-Sys/OFASys}} ~\cite{bai2022}, a unified encoder-decoder model with multi-task scheduling for fine-tuning and additional modality-specific encoders and decoders;
    \item Unified-IO-2\footnote{Accessed 1 Jan 2024: \url{https://github.com/allenai/unified-io-2.pytorch/}}~\cite{lu2023}, a unified encoder-decoder model with additional modality-specific encoders and decoders.
\end{itemize}

All the models encode multimodal data into the same latent representation space as text, pass text and multimodal data jointly to a universal processing unit, and generate output in an autoregressive fashion.
We considered other models such as Macaw~\cite{lyu2023} and OneLLM~\cite{han2023} but did not adopt them due to the higher difficulty of implementation and seemingly poor performance outside of the image domain, even after fine-tuning.

\subsection{Evaluation protocol}
\label{sec:data}

We evaluate the proposed methods for data sampling using the following QA datasets that have free-response questions:

\begin{itemize}
    \item VQA~\cite{antol2015}: questions about MS COCO images depicting everyday life scenes;
    \item iVQA~\cite{yang2021}: questions about how-to videos from YouTube;
    \item Spoken-SQuAD~\cite{lee18}: questions about spoken Wikipedia articles (audio via text-to-speech).
\end{itemize}

We choose these tasks to approximate the kinds of tasks expected of MFMs, seen in the original work on OFASys and Unified-IO~\cite{lu2022,wang2022}.
We use a total of 1500 prompts and associated data (images, audio, video) for each dataset.
Throughout evaluation (\ref{sec:results}), we use the same 80\%/20\% train/test split of the data for fair comparison.
Some of the models in this study have size limitations on the image and video data, which require extra data preprocessing.
For example: for images and videos, we pad and sub-sample the data to 224 x 224 pixels; for videos, we sample 1-second frames for a total of 8 frames per video.

To match the open-response format of the QA tasks, we evaluate the models' performance using typical text generation metrics including BLEU, ROUGE, and BERT-Score~\cite{zhang2020}.
In general, we find that the BERT-Score results have higher values than BLEU and ROUGE, due to the fuzzy nature of the metric (semantic similarity) as compared to the more strict word-overlap metrics that penalize slight changes in wording.

\begin{table*}[t!]
\scriptsize
\centering
\begin{tabular}{l l l l l | l l l}
                       &                  & \multicolumn{3}{c}{OFASys}     & \multicolumn{3}{c}{Unified-IO} \\ \toprule
Modality               & Prompt type            & BLEU     & ROUGE    & BERT     & BLEU     & ROUGE    & BERT     \\ \midrule
\multirow{4}{*}{Audio} & Original         & 0.0434 (0.0052) & 0.1286 (0.0081) & 0.8153 (0.0023) & 0.0142 (0.0037) & \textbf{0.0542} (0.0069) & \textbf{0.8215} (0.0021) \\
                       & Back-translation & \textbf{0.0468} (0.0041) & 0.1316 (0.0061) & 0.8150 (0.00174) & 0.0115 (0.0022) & 0.0437 (0.0038) & 0.8171 (0.0014) \\
                       & Paraphrase       & 0.0431 (0.0029) & 0.1288 (0.0047) & 0.8151 (0.0014) & 0.0113 (0.0015) & 0.0484 (0.0032) & 0.8143 (0.0010) \\
                       & LLM-Paraphrase   & 0.0453 (0.0017) & \textbf{0.1341} (0.0027) & \textbf{0.8168} (0.0008) & 0.0142 (0.0013) & 0.0476 (0.0020) & 0.8172 (0.0006)  \\ \midrule
\multirow{4}{*}{Image} & Original         & \textbf{0.4647} (0.0271) & \textbf{0.4739} (0.0270) & 0.9646 (0.0036) & \textbf{0.6878} (0.0248) & \textbf{0.7012} (0.0245) & \textbf{0.9808} (0.0026) \\
                       & Back-translation & 0.4053 (0.0233) & 0.4145 (0.0233)  & \textbf{0.9648} (0.0031) & 0.5551 (0.0233) & 0.5664 (0.0232) & 0.9743 (0.0026) \\
                       & Paraphrase       & 0.3996 (0.0165) & 0.4112 (0.0165) & 0.9633 (0.0023) & 0.4942 (0.0167) & 0.5051 (0.0167) & 0.9605 (0.0023) \\
                       & LLM-Paraphrase   & 0.3832 (0.0089) & 0.3928 (0.0089) & 0.9636 (0.0012) & 0.3470 (0.0087)  & 0.3563 (0.0087) & 0.9272 (0.0016) \\ \midrule
\multirow{4}{*}{Video} & Original         & \textbf{0.0043} (0.0005)  & \textbf{0.0299} (0.0037)  & 0.8174 (0.0005) & 0.0171 (0.0033) & \textbf{0.0877} (0.0077) & \textbf{0.8294} (0.0016) \\
                       & Back-translation & 0.0038 (0.0004) & 0.0266 (0.0027) & \textbf{0.8177} (0.0004) & 0.0113 (0.0016) & 0.0659 (0.0051) & 0.8281 (0.0012) \\
                       & Paraphrase       & 0.0037 (0.0003) & 0.0259 (0.0021) & 0.8173 (0.0003) & \textbf{0.0193} (0.0023)  & 0.0658 (0.0046) & 0.8257 (0.0010) \\
                       & LLM-Paraphrase   & 0.0039 (0.0002) & 0.0279 (0.0012) & 0.8170 (0.0002) & 0.0058 (0.0004) & 0.0334 (0.0014) & 0.8176 (0.0003)
\end{tabular}
\caption{Performance of baseline models (no additional training) on original and perturbed prompts; standard error in parentheses.}
\label{tab:perturb_acc}
\end{table*}

\section{Results}
\label{sec:results}

\subsection{Testing on perturbed data}
\label{sec:testing_perturbations}

We first assess the accuracy of the baseline MFMs in response to perturbed prompts generated via three separate models:

\begin{enumerate}
    \item General-purpose LLM (see~\ref{sec:data_sampling})
    \item Decoder model fine-tuned on paraphrasing\footnote{Model accessed 30 Aug 2023: \url{https://huggingface.co/ramsrigouthamg/t5_sentence_paraphraser}.}
    \item Back-translation model (between English and Russian; chosen for sufficient linguistic distance to increase distance from original prompt)\footnote{Models accessed 30 Aug 2023: \url{https://huggingface.co/Helsinki-NLP/opus-mt-en-ru} \url{https://huggingface.co/Helsinki-NLP/opus-mt-ru-en}.}
\end{enumerate}

We choose three different methods to assess a wide range of possible prompt changes that end users might make.

The results are shown in Table~\ref{tab:perturb_acc}, in which the image modality has the most consistent drop in performance.
We see consistent degradation of performance from the original questions to the perturbed prompts, up to a 50\% drop in the case of Unified-IO and images (BLEU score of 0.687 for original prompts vs. 0.347 for LLM-paraphrase prompts).
The video domain has slight degradation and low performance overall, due in part to the more complicated answers expected of the model (e.g., ``The device being used in the video is a blender.'') as well as the more complex nature of videos in general.
The overall low performance in the audio domain may be due to the inherently difficult nature of the QA task, which requires first converting speech to text and then performing logical reasoning over the text to answer the question.
Surprisingly, both OFASys and Unified-IO reported high performance in audio and video tasks in the original work (e.g., up to 42.1\% accuracy for OFASys on VQA for MSR-VTT data; \citeauthor{bai2022} \citeyear{bai2022}), which may indicate overfitting on a specific format of the task.
We provide additional results in \cref{sec:modern_model_experiments} using modern models (Phi-4 and Gemma-3n), which support the finding of performance degradation on text data and inconsistent trends in modality-specific data perturbations.

Based on this analysis, we conclude that prompt stability is a major concern for generalist multimodal models even in ``strong'' domains such as images, and a more minor problem for less well-supported domains such as video.

\subsection{Training on perturbed data}
\label{sec:training_perturbations}

\subsubsection{Training conditions}

We address the performance degradation on perturbed data by re-training the MFMs using the prompts generated through data augmentation (\ref{sec:data_sampling}), applied to the QA datasets.
We follow the same training procedure across all models, using the same hyperparameters\footnote{3 training epochs, learning rate $5e-5$, batch size 2, Adam optimizer with default hyperparameters.} and objective function of cross-entropy on the model output logit probabilities compared to the ground-truth data.
For the original prompt condition, we only train the model on the original prompts from all QA datasets.\footnote{We also tried a condition where we trained on the same number of original prompts as the sample prompts, i.e. repeat each original prompt $N=3$ times in the training data, and found substantially worse performance across models and tasks.}
The decoder takes 99.3M parameters for OFASys and 409M parameters for Unified-IO, which is sufficiently lightweight to train on a single A100 GPU.

For each prompt perturbation condition, we use 3 sampled prompts from that condition in place of the original prompt: e.g., for the ``Modality-similarity'' condition, we use the top-3 prompt perturbations based on similarity to the modality data.
OFASys and Unified-IO have a similar architecture that uses an LLM as the decoder for the combined input from text and other modalities, and we only retrain this decoder in both models as we are primarily interested in improving stability in the generated output directly from the decoder, rather than improving stability in e.g. encoder representation.

To summarize, we evaluate the following training conditions for each model type: No training; Train only on original prompt (``Original''); Train on prompt perturbations (``Joint similarity'', ``Modality-similarity'', ``Text similarity'', ``Random sample''; see~\ref{sec:data_sampling}).

\subsubsection{Results}

We first show the model results on the original prompts in Table~\ref{tab:train_model_acc}.
The models without training (``No training'') perform poorly on audio and video and reasonably well on image.
The models trained on the original prompts perform slightly better than the models trained on the perturbed prompts for the image domain, worse for the audio domain, and roughly the same for the video domain.
The models trained on random samples of prompt perturbations generally perform worse than the other perturbation-trained models (e.g., image domain for OFASys), which may be due to noisy prompts that have little semantic relation with the meaning of the original prompt.
In general, the results don't show evidence of \emph{catastrophic forgetting}~\cite{kemker2018}, where the models re-trained on prompt perturbations completely lose the ability to understand the original prompts due to overfitting on the perturbations.

\begin{table*}[]
\centering
\scriptsize
\begin{tabular}{l l l l l | l l l}
                       &                                    & \multicolumn{3}{c}{OFASys}     & \multicolumn{3}{c}{Unified-IO} \\ \toprule
Modality               & Training condition                              & BLEU     & ROUGE    & BERT     & BLEU     & ROUGE    & BERT     \\ \midrule
\multirow{6}{*}{Audio} & No training                        & 0.0434 (0.0052) & 0.1286 (0.0081) & 0.8153 (0.0023) & 0.0142 (0.0037) & 0.0542 (0.0069) & 0.8215 (0.0021) \\
                       & Original prompts                   & 0.1671 (0.0193) & 0.2061 (0.0209) & 0.8882 (0.0043) & 0.0881 (0.0144) & 0.1188 (0.0157) & 0.8629 (0.0037) \\
                       & Joint similarity  & 0.3749 (0.0266) & 0.4034 (0.0267) & 0.9131 (0.0049) & 0.2016 (0.0218) & 0.2248 (0.0222) & 0.8810 (0.0043) \\
                       & Modality-similarity & 0.3708 (0.0266) & 0.3982 (0.0268) & 0.9125 (0.0048) & 0.2205 (0.0227) & 0.2440 (0.0230) & \textbf{0.8883} (0.0044) \\
                       & Text-similarity    & \textbf{0.3910} (0.0269) & \textbf{0.4169} (0.0269) & \textbf{0.9166} (0.0049) & \textbf{0.2303} (0.0231) & \textbf{0.2526} (0.0234) & 0.8857 (0.0045) \\
                       & Random sample                 & 0.3445 (0.0260) & 0.3743 (0.0263) & 0.9095 (0.0048) & 0.2089 (0.0223) & 0.2305 (0.0227) & 0.8813 (0.0044) \\ \midrule
\multirow{6}{*}{Image} & No training                        & 0.4647 (0.0271) & 0.4739 (0.0270) & 0.9646 (0.0036) & 0.6878 (0.0248) & 0.7012 (0.0245) & 0.9808 (0.0026) \\
                       & Original prompts                   & \textbf{0.7102} (0.0243) & \textbf{0.7188} (0.0240) & \textbf{0.9812} (0.0027) & 0.7400 (0.0237) & 0.7507 (0.0233) & 0.9829 (0.0025) \\
                       & Joint similarity  & 0.6650 (0.0255) & 0.6733 (0.0253) & 0.9791 (0.0029) & 0.7186 (0.0241) & 0.7306 (0.0237) & 0.9836 (0.0024) \\
                       & Modality-similarity & 0.6494 (0.0257) & 0.6589 (0.0255) & 0.9767 (0.0030) & 0.7208 (0.0242) & 0.7334 (0.0239) & 0.9834 (0.0025)  \\
                       & Text-similarity    & 0.6463 (0.0258) & 0.6572 (0.0256) & 0.9738 (0.0032) & \textbf{0.7401} (0.0237) & \textbf{0.7510} (0.0234) & \textbf{0.9845} (0.0025) \\
                       & Random sample                 & 0.6189 (0.0260) & 0.6294 (0.0259) & 0.9734 (0.0032) & 0.7130 (0.0243) & 0.7225 (0.0240) & 0.9835 (0.0025) \\ \midrule
\multirow{6}{*}{Video} & No training                        & 0.0043 (0.0005)  & 0.0300 (0.0037)  & 0.8174 (0.0005) & 0.0171 (0.0033) & 0.0877 (0.0077) & 0.8294 (0.0016)  \\
                       & Original prompts                   & 0.4524 (0.0140) & \textbf{0.6499} (0.0116) & 0.9509 (0.0017) & \textbf{0.4927} (0.0160)  & \textbf{0.6703} (0.0130) & \textbf{0.9539} (0.0018) \\
                       & Joint similarity  & 0.4472 (0.0155) & 0.6285 (0.0129) & 0.9485 (0.0019) & 0.4597 (0.0176) & 0.6302 (0.0146) & 0.9467 (0.0021) \\
                       & Modality-similarity & 0.4483 (0.0156) & 0.6286 (0.0128) & 0.9483 (0.0019) & 0.4507 (0.0170) & 0.6231 (0.0142) & 0.9461 (0.0021)  \\
                       & Text-similarity    & \textbf{0.4683} (0.0159) & 0.6474 (0.0127) & \textbf{0.9513} (0.0018) & 0.4914 (0.0183) & 0.6537 (0.0147) & 0.9499 (0.0021) \\
                       & Random sample                 & 0.4387 (0.0152)  & 0.6232 (0.0125) & 0.9478 (0.0018) & 0.4503 (0.0177) & 0.6250 (0.0147) & 0.9464 (0.0021)
\end{tabular}
\caption{Model accuracy on original prompts; standard error in parentheses.}
\label{tab:train_model_acc}
\end{table*}

Next, we evaluate model performance on all prompt perturbations ($\hat{T}$) generated for data augmentation (\ref{sec:data_sampling}), to determine whether a particular form of data-grounded prompt perturbation yield better generalization.
The results are shown in Table~\ref{tab:train_model_perturb_acc}.
Across all modalities, the baseline model and the model trained on original prompts show substantial decrease in performance on prompt perturbations versus the original prompts.
In contrast, the models trained on the prompt perturbations show consistently high performance on the prompt perturbations, with a slight decrease in the video modality and an increase in the audio modality.
In the video domain, the original prompt-trained model has substantially worse performance than the perturbation-trained models, possibly due to a more limited semantic space in the original prompts in the video domain as compared to the prompt perturbations (e.g. perturbations contain variety of verbs to describe the same motion or actions taken in the videos).

Surprisingly, the perturbation-trained models show consistent performance regardless of the prompt sampling method used.
This implies that simply exposing the model to a wider diversity of prompts always helps when testing on unseen perturbations, even if the exact nature of the generated prompts isn't consistent.

\begin{table*}[t!]
\centering
\scriptsize
\begin{tabular}{l l l l l | l l l}
                       &                                    & \multicolumn{3}{c}{OFASys}     & \multicolumn{3}{c}{Unified-IO} \\ \toprule
Modality               & Model                              & BLEU     & ROUGE    & BERT     & BLEU     & ROUGE    & BERT     \\ \midrule
\multirow{6}{*}{Audio} & No training                        & 0.0453 (0.0017) & 0.1341 (0.0027) & 0.8168 (0.0008) & 0.0142 (0.0013) & 0.0476 (0.0020) & 0.8172 (0.0006)  \\
                       & Original prompts                   & 0.1186 (0.0053) & 0.1565 (0.0060)  & 0.8741 (0.0014) & 0.0427 (0.0030) & 0.0715 (0.0036) & 0.8531 (0.0011) \\
                       & Joint similarity  & 0.4367 (0.0088)  & 0.4701 (0.0088) & 0.9208 (0.0016) & 0.2128 (0.0074) & 0.2316 (0.0074) & 0.8820 (0.0014) \\
                       & Modality-similarity & 0.4390 (0.0089) & 0.4714 (0.0088) & 0.9206 (0.0016) & 0.2230 (0.0075) & 0.2415 (0.0076) & 0.8832 (0.0015) \\
                       & Text-similarity    & \textbf{0.4542} (0.0089) & \textbf{0.4860} (0.0088) & \textbf{0.9223} (0.0016) & \textbf{0.2317} (0.0076) & \textbf{0.2507} (0.0077) & \textbf{0.8848} (0.0015) \\
                       & Random sample                 & 0.4090 (0.0087) & 0.4453 (0.0087) & 0.9172 (0.0015) & 0.2311 (0.0076) & 0.2494 (0.0077) & 0.8835 (0.0015) \\ \midrule
\multirow{6}{*}{Image} & No training                        & 0.3832 (0.0089) & 0.3928 (0.0089) & 0.9636 (0.0012) & 0.3470 (0.0087)  & 0.3563 (0.0087) & 0.9272 (0.0016) \\
                       & Original prompts                   & 0.4572 (0.0091) & 0.4647 (0.0091) & 0.9668 (0.0012) & 0.3754 (0.0088) & 0.3851 (0.0088) & 0.9386 (0.0019) \\
                       & Joint similarity  & \textbf{0.6963} (0.0084) & \textbf{0.7002} (0.0084) & \textbf{0.9796} (0.0010) & 0.6371 (0.0088) & 0.6415 (0.0088) & 0.9791 (0.0010) \\
                       & Modality-similarity & 0.6912 (0.0085) & 0.6952 (0.0084) & 0.9793 (0.0010) & 0.6470 (0.0088) & 0.6512 (0.0087) & 0.9800 (0.0009) \\
                       & Text-similarity    & 0.6885 (0.0085) & 0.6941 (0.0084) & 0.9795 (0.0010) & 0.6226 (0.0089) & 0.6280 (0.0088) & 0.9771 (0.00100) \\
                       & Random sample                 & 0.6909 (0.0085) & 0.6961 (0.0084)  & 0.9791 (0.0010)  & \textbf{0.6495} (0.0088) & \textbf{0.6532} (0.0087) & \textbf{0.9801} (0.0009) \\ \midrule
\multirow{6}{*}{Video} & No training                        & 0.0039 (0.0002) & 0.0279 (0.0012) & 0.8170 (0.0002) & 0.0058 (0.0004) & 0.0334 (0.0014) & 0.8176 (0.0003) \\
                       & Original prompts                   & 0.0238 (0.0013) & 0.0730 (0.0025) & 0.8262 (0.0005) & 0.1390 (0.0034) & 0.2720 (0.0049) & 0.8526 (0.0028) \\
                       & Joint similarity  & \textbf{0.3857} (0.0047) & 0.5760 (0.0042) & 0.9414 (0.0006) & 0.4301 (0.0063) & 0.5878 (0.0052) & 0.9417 (0.0008) \\
                       & Modality-similarity & 0.3824 (0.0047) & 0.5752 (0.0041) & 0.9412 (0.0006) & 0.4394 (0.0065) & \textbf{0.5947} (0.0053) & 0.9422 (0.0008) \\
                       & Text-similarity    & 0.3854 (0.0047) & \textbf{0.5765} (0.0041) & \textbf{0.9416} (0.0006) & 0.4378 (0.0064)  & 0.5937 (0.0053) & 0.9424 (0.0008) \\
                       & Random sample                 & 0.3825 (0.0047) & 0.5744 (0.0041) & 0.9413 (0.0006) & \textbf{0.4401} (0.0064) & 0.5936 (0.0053) & \textbf{0.9426} (0.0008)
\end{tabular}
\caption{Model accuracy on prompt perturbations; standard error in parentheses.}
\label{tab:train_model_perturb_acc}
\end{table*}

\begin{table*}[t!]
\small
\centering
\begin{tabular}{lllll | lll}
                       &                                    & \multicolumn{3}{c}{OFASys}     & \multicolumn{3}{c}{Unified-IO} \\ \toprule
Modality               & Model                              & BLEU     & ROUGE    & BERT     & BLEU     & ROUGE    & BERT     \\ \midrule
\multirow{6}{*}{Audio} & No training                        & 2.0588 & 1.0597 & \textbf{0.0506} & 5.0223 & 2.2407 & \textbf{0.0401} \\
                       & Original prompts                   & 2.3907 & 2.0382 & 0.0838 & 3.8096 & 2.7091 & 0.0691  \\
                       & Joint similarity  & 1.0825 & 1.0023 & 0.0908 & 1.860  & 1.7236 & 0.0874  \\
                       & Modality-similarity & 1.0811 & 1.0024 & 0.0905 & 1.8092 & 1.6807 & 0.0883 \\
                       & Text-similarity    & \textbf{1.0484} & \textbf{0.9740} & 0.0901 & \textbf{1.7637} & \textbf{1.6400} & 0.0882 \\
                       & Random sample               & 1.1384 & 1.0469 & 0.0900 & 1.7672 & 1.6517   & 0.0927 \\ \midrule
\multirow{6}{*}{Image} & No training                        & 1.2503 & 1.2221 & 0.0689 & 1.3487 & 1.3197 & 0.0930 \\
                       & Original prompts                   & 1.0750 & 1.0576 & 0.0659 & 1.2685 & 1.2372 & 0.1103 \\
                       & Joint similarity  & \textbf{0.6543} & \textbf{0.6473} & \textbf{0.0545} & 0.7499 & 0.7413 & 0.0533 \\
                       & Modality-similarity & 0.6630 & 0.6558 & 0.0549 & 0.7333 & 0.7255 & 0.0521 \\
                       & Text-similarity    & 0.6657 & 0.6551  & 0.0550 & 0.7708 & 0.7607 & 0.0555 \\
                       & Random sample               & 0.6632 & 0.6536 & 0.0552 & \textbf{0.7300} & \textbf{0.7231} & \textbf{0.0519} \\ \midrule
\multirow{6}{*}{Video} & No training                        & 2.2890 & 2.2632 & \textbf{0.0103} & 3.7975 & 2.3034 & \textbf{0.0183} \\
                       & Original prompts                   & 3.0050 & 1.8506 & 0.0354 & 1.3150 & 0.9801 & 0.1785  \\
                       & Joint similarity  & 0.6646 & \textbf{0.3932} & 0.0348 & 0.7952 & 0.4805 & 0.0437  \\
                       & Modality-similarity & \textbf{0.6595} & 0.3842 & 0.0345 & 0.7952 & 0.4815 & 0.0446 \\
                       & Text-similarity    & 0.6623 & 0.3890  & 0.0347 & 0.7906 & 0.4794 & 0.0441   \\
                       & Random sample               & 0.6629   & 0.3909 & 0.0348 & \textbf{0.7853} & \textbf{0.4817} & 0.0441
\end{tabular}
\caption{Model coefficient of variation, on prompt perturbations (lower is better).}
\label{tab:train_model_perturb_var}
\end{table*}

To assess the consistency of the performance differences over the prompt perturbations test data, we compute the coefficient of variation $CV$ ($\frac{\sigma}{\mu}$) over all the prompt perturbations.
Using the performance scores for all output from a given model and task, we compute the standard deviation in scores, and normalize the standard deviation of the scores by the mean.
The normalized value represents the amount of variation in score relative to the expected score, e.g. a standard deviation of 5 in BLEU score for a high-scoring model should be better than the same standard deviation for a low-scoring model.

Shown in Table~\ref{tab:train_model_perturb_var}, the re-trained models achieve lower $CV$ on the prompt perturbations.
This again confirms that training with any of the prompt perturbation strategies yields more stable performance as compared to training only on the original prompts.
The variation is slightly higher in the audio domain, which aligns with the generally lower performance on all forms of prompt perturbations in audio (Table~\ref{tab:perturb_acc}), such that even a small absolute error is relatively large in comparison to the overall lower performance.




\begin{table*}[t!]
\centering
\scriptsize
\begin{tabular}{l L{1cm} L{2cm} L{1.25cm} L{1.25cm} L{1.25cm} | L{1cm} L{1cm} L{1cm} L{1cm}} 
& & & & & & \multicolumn{2}{c}{OFASys} & \multicolumn{2}{c}{Unified-IO} \\ 
Modality & Cluster theme & Example data & Prompt & Perturb prompt & True answer & Joint similarity & Original prompts & Joint similarity & Original prompts \\ \toprule
Audio & Religion & Spoken: ``In the United States, the UMC ranks as the largest mainline Protestant denomination, the largest Protestant church after the Southern Baptist Convention, and the third largest Christian denomination.'' & In the United States the UMC ranks as the largest what? & in the united states, the umc ranks as the largest \_\_\_\_\_\_\_ of christian organizations. & mainline protestant denomination & largest protestant & christian & mainline protestant denomination & national \\ \hline
Image & Home activities & Well-lit kitchen with a table, preparation station, counters, cabinets, oven, and sink. & What is the table made of? & what does the table primarily consist of in terms of substance? & marble & marble & glass & marble & table \\ \hline
Video & Personal grooming & Person with long hair sitting in a room, brushing their hair. & How was the girl combing her hair? & how did she groom her hair? & The girl in the video was combing her hair with a brush. & The woman in the video combed her hair with a brush. & dry & The girl in the video was brushing her hair with a brush. & she brushed her hair \\
\end{tabular}
\caption{Example data from target modalities, clustered by semantic representation. Cluster themes determined by manual inspection of prompts. Models organized by architecture and by training condition: ``Original prompts'' indicates the model trained only with original prompts, ``Joint similarity'' indicates the model trained on prompt perturbations filtered for similarity to modality data and text data.}
\label{tab:data_cluster_example_output}
\end{table*}

\subsection{Error analysis}
\label{sec:error_analysis}

We perform a systematic error analysis to determine possible reasons for the performance improvement on the prompt perturbations (accuracy shown in Table~\ref{tab:train_model_perturb_acc}), and we compare the perturbation-trained models to the prompt-trained models.
We first embed all the modality data using the same ImageBind model used for prompt perturbation filtering (\ref{sec:data_sampling}), reduce to $D=3$ dimensions with PCA to group dimensions with similar semantics, and then cluster data from each modality via cosine similarity of the reduced embeddings.
The clustering is implemented with a hierarchical DBSCAN with default hyperparameters (minimum cluster size 5, automatic algorithm selection) which identified $K={15,11,12}$ clusters for audio, image, and video modalities respectively.
We identify the dominant concept in each cluster based on the associated prompts, and identify clusters with an especially high improvement from the perturbation-trained models as compared to the prompt-trained models.

Examples of high-improvement clusters are shown in Table~\ref{tab:data_cluster_example_output}.
For audio, the example cluster represents spoken explanations of religious organizations, such as the UMC (United Methodist Church).
This topic of data likely requires more complicated and contextualized answers to questions, which the perturbation-trained models seem able to provide (e.g. OFASys response, ``largest protestant'') while the models trained only on original prompts produce more short and ambiguous responses (OFASys, ``national'').
For image, the example cluster contains scenes of home activities, including a typical kitchen scene.
These data include prompts that ask about basic object properties such as quantity and material type, and the perturbation-trained models appear more able to understand the intent of the prompts as they relate to properties such as ``substance'' (e.g., Unified-IO, ``marble'').
For video, the example cluster includes clips of people facing the camera, mainly performing tasks with tools such as applying makeup and combing their hair.
The prompts for this data ask for further descriptions of the actions shown, and the prompt-trained models have difficulty understanding the full context of differently-phrased actions (e.g., OFASys, ``dry'') as compared to the perturbation-trained models that provide a complete description of the actions (OFASys, ``The woman in the video combed her hair with a brush.'').

We show more examples of per-cluster prompts and quantitative differences in model performance in Appendix, e.g., the image cluster related to street views shows a roughly 250\% improvement from prompt-trained to perturbation-trained (0.3713 to 0.9050).







\section{Conclusion}

This study showed that multimodal foundation models (MFMs) show a surprising degree of instability in the image domain in response to prompt perturbations (\ref{sec:testing_perturbations}).
However, the instability can be reduced through fine-tuning on additional prompt perturbations, regardless of whether the perturbations are close in semantic space to the original text data or modality data (\ref{sec:training_perturbations}).
The finding about re-training on different sub-sets of prompt perturbations is surprising but suggests that MFM developers should focus more on diversity of prompts rather than optimizing a specific method for prompt sampling, because the choice of method appears less relevant for downstream model performance.
Lastly, the error analysis shows systematic improvements in specific clusters of data in different modalities, e.g. in videos that depict personal grooming (\ref{sec:error_analysis}).
This suggests that training on prompt perturbations helps reinforce modality-specific properties (e.g., fine-grained visual details in grooming behavior), that in turn leads to greater robustness in those modalities.

This study provides just one approach to tackle the larger problem of model robustness, which future work should continue to address with different methods.
For one, future studies may investigate more model-driven paradigms for increasing robustness, including direct preference optimization to determine the relative value of the different perturbations for model training~\cite{rafailov2024,zhao2024}.
In this work we only considered perturbations of the text data, but MFMs should also be tested with respect to perturbations of the modality data, e.g., adding additional noise to images~\cite{lee2020}.
Lastly, future work should consider more structured or explainable methods for determining similarity between modality data and text.
E.g., one might use the text output from a well-trained image caption model to compare against a prompt perturbation to determine if the perturbation has ``drifted'' too far from the image content.

\section{Acknowledgments}

This work was supported by the NNSA Office of Defense Nuclear Nonproliferation Research and Development, U.S. Department of Energy, and Pacific Northwest National Laboratory, which is operated by Battelle Memorial Institute for the U.S. Department of Energy under Contract DEAC05–76RLO1830.
This article has been cleared by PNNL for public release as PNNL-SA-201679.

\bibliography{main}

@inproceedings{wang2022,
  title={{OFA: Unifying architectures, tasks, and modalities through a simple sequence-to-sequence learning framework}},
  author={Wang, Peng and Yang, An and Men, Rui and Lin, Junyang and Bai, Shuai and Li, Zhikang and Ma, Jianxin and Zhou, Chang and Zhou, Jingren and Yang, Hongxia},
  booktitle={International Conference on Machine Learning},
  pages={23318--23340},
  year={2022},
  organization={PMLR}
}

@article{bai2022,
  title={{OFASys: A multi-modal multi-task learning system for building generalist models}},
  author={Bai, Jinze and Men, Rui and Yang, Hao and Ren, Xuancheng and Dang, Kai and Zhang, Yichang and Zhou, Xiaohuan and Wang, Peng and Tan, Sinan and Yang, An and others},
  journal={arXiv preprint arXiv:2212.04408},
  year={2022}
}

@article{lyu2023,
  title={{Macaw-LLM}: Multi-modal language modeling with image, audio, video, and text integration},
  author={Lyu, Chenyang and Wu, Minghao and Wang, Longyue and Huang, Xinting and Liu, Bingshuai and Du, Zefeng and Shi, Shuming and Tu, Zhaopeng},
  journal={arXiv preprint arXiv:2306.09093},
  year={2023}
}

@inproceedings{kemker2018,
  title={Measuring catastrophic forgetting in neural networks},
  author={Kemker, Ronald and McClure, Marc and Abitino, Angelina and Hayes, Tyler and Kanan, Christopher},
  booktitle={Proceedings of the AAAI conference on artificial intelligence},
  volume={32},
  number={1},
  year={2018}
}

@article{wu2023,
  title={{Next-GPT: Any-to-any multimodal LLM}},
  author={Wu, Shengqiong and Fei, Hao and Qu, Leigang and Ji, Wei and Chua, Tat-Seng},
  journal={arXiv preprint arXiv:2309.05519},
  year={2023}
}

@inproceedings{lee2020,
  title={Smoothmix: a simple yet effective data augmentation to train robust classifiers},
  author={Lee, Jin-Ha and Zaheer, Muhammad Zaigham and Astrid, Marcella and Lee, Seung-Ik},
  booktitle={Proceedings of the IEEE/CVF conference on computer vision and pattern recognition workshops},
  pages={756--757},
  year={2020}
}

@inproceedings{zhang2020,
  title={{BERTScore: Evaluating Text Generation with BERT}},
  author={Zhang, Tianyi and Kishore, Varsha and Wu, Felix and Weinberger, Kilian Q and Artzi, Yoav},
  booktitle={International Conference on Learning Representations},
  year={2020}
}

@article{rafailov2024,
  title={Direct preference optimization: Your language model is secretly a reward model},
  author={Rafailov, Rafael and Sharma, Archit and Mitchell, Eric and Manning, Christopher D and Ermon, Stefano and Finn, Chelsea},
  journal={Advances in Neural Information Processing Systems},
  volume={36},
  year={2024}
}

@inproceedings{zhao2024,
    title = "Improving the Robustness of Large Language Models via Consistency Alignment",
    author = "Zhao, Yukun  and
      Yan, Lingyong  and
      Sun, Weiwei  and
      Xing, Guoliang  and
      Wang, Shuaiqiang  and
      Meng, Chong  and
      Cheng, Zhicong  and
      Ren, Zhaochun  and
      Yin, Dawei",
    editor = "Calzolari, Nicoletta  and
      Kan, Min-Yen  and
      Hoste, Veronique  and
      Lenci, Alessandro  and
      Sakti, Sakriani  and
      Xue, Nianwen",
    booktitle = "Proceedings of the 2024 Joint International Conference on Computational Linguistics, Language Resources and Evaluation (LREC-COLING 2024)",
    month = may,
    year = "2024",
    address = "Torino, Italia",
    publisher = "ELRA and ICCL",
    url = "https://aclanthology.org/2024.lrec-main.782",
    pages = "8931--8941",
}

@article{han2023,
  title={{OneLLM: One framework to align all modalities with language}},
  author={Han, Jiaming and Gong, Kaixiong and Zhang, Yiyuan and Wang, Jiaqi and Zhang, Kaipeng and Lin, Dahua and Qiao, Yu and Gao, Peng and Yue, Xiangyu},
  journal={arXiv preprint arXiv:2312.03700},
  year={2023}
}

@inproceedings{cui2024,
  title={A survey on multimodal large language models for autonomous driving},
  author={Cui, Can and Ma, Yunsheng and Cao, Xu and Ye, Wenqian and Zhou, Yang and Liang, Kaizhao and Chen, Jintai and Lu, Juanwu and Yang, Zichong and Liao, Kuei-Da and others},
  booktitle={Proceedings of the IEEE/CVF Winter Conference on Applications of Computer Vision},
  pages={958--979},
  year={2024}
}

@article{moor2023,
  title={Foundation models for generalist medical artificial intelligence},
  author={Moor, Michael and Banerjee, Oishi and Abad, Zahra Shakeri Hossein and Krumholz, Harlan M and Leskovec, Jure and Topol, Eric J and Rajpurkar, Pranav},
  journal={Nature},
  volume={616},
  number={7956},
  pages={259--265},
  year={2023},
  publisher={Nature Publishing Group UK London}
}

@inproceedings{lu2022,
  title={{Unified-IO: A unified model for vision, language, and multi-modal tasks}},
  author={Lu, Jiasen and Clark, Christopher and Zellers, Rowan and Mottaghi, Roozbeh and Kembhavi, Aniruddha},
  booktitle={The Eleventh International Conference on Learning Representations},
  year={2022}
}

@article{lu2023,
  title={Unified-io 2: Scaling autoregressive multimodal models with vision, language, audio, and action},
  author={Lu, Jiasen and Clark, Christopher and Lee, Sangho and Zhang, Zichen and Khosla, Savya and Marten, Ryan and Hoiem, Derek and Kembhavi, Aniruddha},
  journal={arXiv preprint arXiv:2312.17172},
  year={2023}
}

@inproceedings{khattak2023,
  title={{MaPLe: Multi-modal prompt learning}},
  author={Khattak, Muhammad Uzair and Rasheed, Hanoona and Maaz, Muhammad and Khan, Salman and Khan, Fahad Shahbaz},
  booktitle={Proceedings of the IEEE/CVF Conference on Computer Vision and Pattern Recognition},
  pages={19113--19122},
  year={2023}
}

@inproceedings{lee2023,
  title={Multimodal prompting with missing modalities for visual recognition},
  author={Lee, Yi-Lun and Tsai, Yi-Hsuan and Chiu, Wei-Chen and Lee, Chen-Yu},
  booktitle={Proceedings of the IEEE/CVF Conference on Computer Vision and Pattern Recognition},
  pages={14943--14952},
  year={2023}
}

@article{li2024,
  title={Multimodal foundation models: From specialists to general-purpose assistants},
  author={Li, Chunyuan and Gan, Zhe and Yang, Zhengyuan and Yang, Jianwei and Li, Linjie and Wang, Lijuan and Gao, Jianfeng and others},
  journal={Foundations and Trends{\textregistered} in Computer Graphics and Vision},
  volume={16},
  number={1-2},
  pages={1--214},
  year={2024},
  publisher={Now Publishers, Inc.}
}

@inproceedings{li2023,
    title = "Towards Robust Pruning: An Adaptive Knowledge-Retention Pruning Strategy for Language Models",
    author = "Li, Jianwei  and
      Lei, Qi  and
      Cheng, Wei  and
      Xu, Dongkuan",
    editor = "Bouamor, Houda  and
      Pino, Juan  and
      Bali, Kalika",
    booktitle = "Proceedings of the 2023 Conference on Empirical Methods in Natural Language Processing",
    month = dec,
    year = "2023",
    address = "Singapore",
    publisher = "Association for Computational Linguistics",
    url = "https://aclanthology.org/2023.emnlp-main.79",
    doi = "10.18653/v1/2023.emnlp-main.79",
    pages = "1229--1247",
}

@inproceedings{ge2023,
  title={Improving zero-shot generalization and robustness of multi-modal models},
  author={Ge, Yunhao and Ren, Jie and Gallagher, Andrew and Wang, Yuxiao and Yang, Ming-Hsuan and Adam, Hartwig and Itti, Laurent and Lakshminarayanan, Balaji and Zhao, Jiaping},
  booktitle={Proceedings of the IEEE/CVF Conference on Computer Vision and Pattern Recognition},
  pages={11093--11101},
  year={2023}
}

@inproceedings{newman2021,
  title={P-Adapters: Robustly Extracting Factual Information from Language Models with Diverse Prompts},
  author={Newman, Benjamin and Choubey, Prafulla Kumar and Rajani, Nazneen},
  booktitle={International Conference on Learning Representations},
  year={2021}
}

@article{chen2023,
  title={{An empirical survey of data augmentation for limited data learning in NLP}},
  author={Chen, Jiaao and Tam, Derek and Raffel, Colin and Bansal, Mohit and Yang, Diyi},
  journal={Transactions of the Association for Computational Linguistics},
  volume={11},
  pages={191--211},
  year={2023},
  publisher={MIT Press One Broadway, 12th Floor, Cambridge, Massachusetts 02142, USA~…}
}

@inproceedings{schulhoff2023,
  title={{Ignore This Title and HackAPrompt: Exposing Systemic Vulnerabilities of LLMs Through a Global Prompt Hacking Competition}},
  author={Schulhoff, Sander and Pinto, Jeremy and Khan, Anaum and Bouchard, Louis-Fran{\c{c}}ois and Si, Chenglei and Anati, Svetlina and Tagliabue, Valen and Kost, Anson and Carnahan, Christopher and Boyd-Graber, Jordan},
  booktitle={Proceedings of the 2023 Conference on Empirical Methods in Natural Language Processing},
  pages={4945--4977},
  year={2023}
}

@inproceedings{kim2024,
  title={{Understanding users’ dissatisfaction with ChatGPT responses: Types, resolving tactics, and the effect of knowledge level}},
  author={Kim, Yoonsu and Lee, Jueon and Kim, Seoyoung and Park, Jaehyuk and Kim, Juho},
  booktitle={Proceedings of the 29th International Conference on Intelligent User Interfaces},
  pages={385--404},
  year={2024}
}

@inproceedings{formento2023,
  title={Using punctuation as an adversarial attack on deep learning-based NLP systems: An empirical study},
  author={Formento, Brian and Foo, Chuan Sheng and Tuan, Luu Anh and Ng, See Kiong},
  booktitle={Findings of the Association for Computational Linguistics: EACL 2023},
  pages={1--34},
  year={2023}
}

@inproceedings{girdhar2023,
  title={Imagebind: One embedding space to bind them all},
  author={Girdhar, Rohit and El-Nouby, Alaaeldin and Liu, Zhuang and Singh, Mannat and Alwala, Kalyan Vasudev and Joulin, Armand and Misra, Ishan},
  booktitle={Proceedings of the IEEE/CVF Conference on Computer Vision and Pattern Recognition},
  pages={15180--15190},
  year={2023}
}

@inproceedings{antol2015,
author = {Stanislaw Antol and Aishwarya Agrawal and Jiasen Lu and Margaret Mitchell and Dhruv Batra and C. Lawrence Zitnick and Devi Parikh},
title = {{VQA}: {V}isual {Q}uestion {A}nswering},
booktitle = {International Conference on Computer Vision},
year = {2015},
}

@article{team2025gemma,
  title={Gemma 3 technical report},
  author={Kamath, Aishwarya and Ferret, Johan and Pathak, Shreya and Vieillard, Nino and Merhej, Ramona and Perrin, Sarah and Matejovicova, Tatiana and Ram{\'e}, Alexandre and Rivi{\`e}re, Morgane and others},
  journal={arXiv preprint arXiv:2503.19786},
  year={2025}
}

@article{abouelenin2025phi,
  title={Phi-4-mini technical report: Compact yet powerful multimodal language models via mixture-of-loras},
  author={Abouelenin, Abdelrahman and Ashfaq, Atabak and Atkinson, Adam and Awadalla, Hany and Bach, Nguyen and Bao, Jianmin and Benhaim, Alon and Cai, Martin and Chaudhary, Vishrav and Chen, Congcong and others},
  journal={arXiv preprint arXiv:2503.01743},
  year={2025}
}

@inproceedings{yang2021,
  title={Just ask: Learning to answer questions from millions of narrated videos},
  author={Yang, Antoine and Miech, Antoine and Sivic, Josef and Laptev, Ivan and Schmid, Cordelia},
  booktitle={Proceedings of the IEEE/CVF international conference on computer vision},
  pages={1686--1697},
  year={2021}
}

@inproceedings{lee18,
  author={Chia-Hsuan Lee and Szu-Lin Wu and Chi-Liang Liu and Hung-yi Lee},
  title={{Spoken SQuAD: A Study of Mitigating the Impact of Speech Recognition Errors on Listening Comprehension}},
  year=2018,
  booktitle={Proc. Interspeech 2018},
  pages={3459--3463},
  doi={10.21437/Interspeech.2018-1714}
}

\appendix

\section{Experiments on modern multimodal models}
\label{sec:modern_model_experiments}

We conduct additional experiments to validate the perturbation-related results on modern multimodal models (cf. \cref{sec:testing_perturbations}).
The following modern models are tested:

\begin{itemize}
    \item Phi-4~\cite{abouelenin2025phi};
    \item Gemma-3n~\cite{team2025gemma}.
\end{itemize}

We run the same prompt perturbations through the other models and largely replicate the earlier results in which model performance decreases when faced with text perturbations~\cref{sec:modern_model_prompt_perturbations}.
We then perturb the modality-specific data by adding noise, swapping sub-sections of data, and masking sub-sections of data~\cref{sec:modern_model_modality_perturbations}, and we find that these perturbations do not consistently decrease or increase performance, which suggests that the main problem with multimodal model instability relates to text-related perturbations.

\subsection{Prompt perturbations}
\label{sec:modern_model_prompt_perturbations}

We show the results of the prompt perturbations in \autoref{tab:perturb_acc_new_models}.
As seen before, the models' performance on prompt perturbations declines significantly regardless of modality, particularly in the video modality where BLEU scores drop by up to 20\% for Gemma and up to 28\% for Phi-4.
Interestingly, for some perturbations in the image modality for Phi-4 we see increased performance.
We attribute this difference in results to a qualitative change in Phi-4 behavior as a result of receiving perturbed prompts.
Normally Phi-4 output is verbose and contains superfluous information about the data provided, but when providing the model with noisy data it appears that the responses are often more concise, resulting in higher text overlap scores.
Why this only impacts images may be an artifact of the training for Phi-4 that prioritized long detailed captions on images as opposed to other modalities where the fine-tuning may have involved more concise output.

\begin{table*}[t!]
\centering
\scriptsize
\begin{tabular}{l l | l l l | l l l}
                       &                                      & \multicolumn{3}{c}{Gemma-3n}               & \multicolumn{3}{| c}{Phi-4}  \\ \toprule
Modality               & Prompt & BLEU              & ROUGE             & BERT              & BLEU              & ROUGE             & BERT              \\ \midrule
Audio & Original                             & \textbf{0.08460} (0.00520) & \textbf{0.18980} (0.00938) & \textbf{0.82617} (0.00185) & \textbf{0.21653} (0.01333) & \textbf{0.34213} (0.01566) & \textbf{0.86562} (0.00302) \\
                       & LLM-Paraphrase                    & 0.05259 (0.00125) & 0.11038 (0.00242) & 0.82056 (0.00054) & 0.11954 (0.00321) & 0.19136 (0.00403) & 0.84177 (0.00082) \\
                       & Back-translation & 0.06959 (0.00332) & 0.15202 (0.00637) & 0.82510 (0.00125) & 0.13939 (0.00738) & 0.22462 (0.00963) & 0.85045 (0.00185) \\
                       & Paraphrase        & 0.06521 (0.00280) & 0.14177 (0.00534) & 0.82206 (0.00107) & 0.15964 (0.00765) & 0.24218 (0.00866) & 0.84984 (0.00173) \\ \midrule
Image & Original                             & \textbf{0.03542} (0.00244) & \textbf{0.08650} (0.00522) & \textbf{0.81276} (0.00098) & 0.36079 (0.02548) & 0.39153 (0.02451) & 0.90119 (0.00478) \\
                       & LLM-Paraphrase                    & 0.02649 (0.00072) & 0.06486 (0.00151) & 0.81270 (0.00030) & 0.34425 (0.00840) & 0.35425 (0.00826) & 0.91033 (0.00161) \\
                       & Back-translation & 0.03180 (0.00205) & 0.07802 (0.00421) & 0.81208 (0.00081) & \textbf{0.40713} (0.02290) & \textbf{0.42287} (0.02230) & \textbf{0.91883} (0.00413) \\
                       & Paraphrase        & 0.03189 (0.00150) & 0.07864 (0.00314) & 0.81263 (0.00058) & 0.38722 (0.01598) & 0.39820 (0.01560) & 0.91072 (0.00297) \\ \midrule
Video & Original                             & \textbf{0.37934} (0.00659) & \textbf{0.45289} (0.00900) & \textbf{0.89921} (0.00119) & \textbf{0.56810} (0.01209) & \textbf{0.46234} (0.01046) & \textbf{0.92176} (0.00158) \\
                       & LLM-Paraphrase                    & 0.31094 (0.00189) & 0.35088 (0.00250) & 0.88856 (0.00036) & 0.40677 (0.00375) & 0.33885 (0.00308) & 0.90074 (0.00062) \\
                       & Back-translation & 0.32404 (0.00549) & 0.35879 (0.00695) & 0.88985 (0.00100) & 0.42460 (0.00960) & 0.33502 (0.00764) & 0.90196 (0.00138) \\
                       & Paraphrase        & 0.34445 (0.00399) & 0.39269 (0.00534) & 0.89255 (0.00078) & 0.46408 (0.00738) & 0.38279 (0.00619) & 0.90795 (0.00113)
\end{tabular}
\caption{Performance of baseline models (no additional training) on original and perturbed prompts; standard error in parentheses.}
\label{tab:perturb_acc_new_models}
\end{table*}

\subsection{Modality perturbations}
\label{sec:modern_model_modality_perturbations}

A natural question from the prompt perturbation experiments would be whether the same effect holds for perturbations on the modality-specific data, e.g. image noising.
We experiment with three different types of modality perturbations:

\begin{itemize}
    \item Mask: randomly mask 10\% of the data, in regularly-shaped 2-D squares (for image and video) or in 1-D lines (for audio);
    \item Noise: randomly add Gaussian noise to the data (normalized to have the same mean and variance as original data);
    \item Swap: randomly swap 10\% of the data, in regularly-shaped 2-D squares (for image and video) or in 1-D lines (for audio).
\end{itemize}

We run the evaluation on a smaller sample of data due to memory constraints (N=1000 per modality).
The results are shown in \autoref{tab:perturb_modality_acc_new_models}, and we see largely similar performance across modalities regardless of perturbation condition.
The primary exception is the improved performance on modality perturbations in the image modality for Phi-4, as seen in the prompt perturbation results.
In nearly all cases, we can conclude that modality data perturbations pose much less of a problem to multimodal model performance as compared to prompt perturbations.

\begin{table*}[t!]
\centering
\scriptsize
\begin{tabular}{l l | l l l | l l l}
                       &                       & \multicolumn{3}{c}{Gemma-3n}               & \multicolumn{3}{| c}{Phi-4}  \\ \toprule
Modality               & Perturbation  & BLEU              & ROUGE             & BERT              & BLEU              & ROUGE             & BERT              \\ \midrule
Audio & Original              & 0.08568 (0.00480) & 0.19375 (0.00903) & \textbf{0.82850} (0.00178) & \textbf{0.21653} (0.01333) & 0.34213 (0.01566) & \textbf{0.86562} (0.00302) \\
                       & Mask                  & 0.08670 (0.00508) & 0.19385 (0.00928) & 0.82680 (0.00177) & 0.20163 (0.01144) & 0.32878 (0.01450) & 0.86228 (0.00280) \\
                       & Noise                 & \textbf{0.08864} (0.00520) & \textbf{0.19667} (0.00966) & 0.82759 (0.00187) & 0.21017 (0.01217) & \textbf{0.34339} (0.01566) & 0.86528 (0.00289) \\
                       & Swap                  & 0.08132 (0.00501) & 0.18277 (0.00966) & 0.82553 (0.00190) & 0.19848 (0.01079) & 0.33010 (0.01483) & 0.86020 (0.00273) \\ \midrule
Image & Original              & \textbf{0.03622} (0.00248) & \textbf{0.08974} (0.00502) & 0.81352 (0.00095) & 0.36079 (0.02548) & 0.39153 (0.02451) & 0.90119 (0.00478) \\
                       & Mask                  & 0.03446 (0.00236) & 0.08462 (0.00506) & 0.81289 (0.00094) & 0.63760 (0.02723) & 0.63353 (0.02710) & 0.96650 (0.00378) \\
                       & Noise                 & 0.03534 (0.00228) & 0.08466 (0.00486) & \textbf{0.81361} (0.00092) & 0.60103 (0.02767) & 0.59879 (0.02741) & 0.95801 (0.00416) \\
                       & Swap                  & 0.03342 (0.00230) & 0.08149 (0.00499) & 0.81341 (0.00094) & \textbf{0.64869} (0.02698) & \textbf{0.64401} (0.02683) & \textbf{0.96906} (0.00373) \\ \midrule
Video & Original              & 0.37991 (0.00657) & 0.44466 (0.00926) & 0.89774 (0.00121) & 0.56810 (0.01209) & 0.46234 (0.01046) & 0.92176 (0.00158) \\
                       & Mask                  & 0.37495 (0.00681) & 0.44338 (0.00918) & \textbf{0.89778} (0.00124) & 0.57727 (0.01168) & 0.46501 (0.01025) & 0.92336 (0.00150) \\
                       & Noise                 & 0.37061 (0.00643) & 0.44293 (0.00907) & 0.89700 (0.00118) & \textbf{0.58019} (0.01263) & \textbf{0.46943} (0.01065) & 0.92325 (0.00162) \\
                       & Swap                  & \textbf{0.38795} (0.00639) & \textbf{0.46090} (0.00911) & 0.89698 (0.00127) & 0.57764 (0.01186) & 0.46758 (0.01031) & \textbf{0.92373} (0.00145) 
\end{tabular}
\caption{Performance of baseline models (no additional training) on original and perturbed modality data; standard error in parentheses.}
\label{tab:perturb_modality_acc_new_models}
\end{table*}

\end{document}